\documentclass[final]{cvpr}

\usepackage{times}
\usepackage{epsfig}
\usepackage{graphicx}
\usepackage{amsmath}
\usepackage{amssymb}
\usepackage{subfigure}
\usepackage{subcaption}
\usepackage{float}
\usepackage{makecell}

\usepackage{CJKutf8}

\usepackage[toc,page]{appendix}

\usepackage{hyperref}
\hypersetup{colorlinks}

\def\confYear{CVPR 2024}

\begin{document}

\title{GradeADreamer: Enhanced Text-to-3D Generation \\ Using Gaussian Splatting and Multi-View Diffusion}

\author{Trapoom Ukarapol \\
Tsinghua University\\
{\tt\small ukarapolt10@mails.tsinghua.edu.cn}
\and
Kevin Pruvost \\
Tsinghua University\\
{\tt\small pkw23@mails.tsinghua.edu.cn}
}

\maketitle

\begin{abstract}

Text-to-3D generation has shown promising results, yet common challenges such as the Multi-face Janus problem and extended generation time for high-quality assets. In this paper, we address these issues by introducing a novel three-stage training pipeline called GradeADreamer. This pipeline is capable of producing high-quality assets with a total generation time of under 30 minutes using only a single RTX 3090 GPU. Our proposed method employs a Multi-view Diffusion Model, MVDream, to generate Gaussian Splats as a prior, followed by refining geometry and texture using StableDiffusion. Experimental results demonstrate that our approach significantly mitigates the Multi-face Janus problem and achieves the highest average user preference ranking compared to previous state-of-the-art methods. The project code is available at \url{https://github.com/trapoom555/GradeADreamer}.

\end{abstract}

\section{Introduction}

3D Content Creation is pivotal in numerous fields, including virtual and augmented reality, game design, head modeling, and beyond. One particularly innovative area is text-to-3D generation, which enables the creation of imaginative 3D models based on textual descriptions. This has emerged as a rapidly growing research area in recent years. Considerable advancements were achieved through deep learning techniques \cite{wu2017learning, chen2018text2shape, 10004645} in the beginning. \\

Currently, more novel methods \cite{chen2023fantasia3d,lin2023magic3d,shi2023MVDream,wang2023prolificdreamer,jain2021dreamfields, khalid2022clipmesh, poole2022dreamfusion, metzer2022latent, ruiz2022dreambooth, zhang2024exactdreamer, chen2024textto3d} utilize 3D representations, such as NeRF \cite{mildenhall2020nerf} or Gaussian Splatting \cite{kerbl3Dgaussians}, for 3D asset representation. These methods optimize the process by distilling a pre-trained text-to-image generation model to guide 3D content generation. 
Papers such as Dreamfusion \cite{poole2022dreamfusion} and Magic3D \cite{lin2023magic3d} employ StableDiffusion \cite{Rombach_2022_CVPR} which use it as supervision for the optimization process of the 3D Model via Score Distillation Sampling (SDS) \cite{poole2022dreamfusion}. \\

\begin{figure}[ht]
    \centering
    \includegraphics[width=0.45\textwidth]{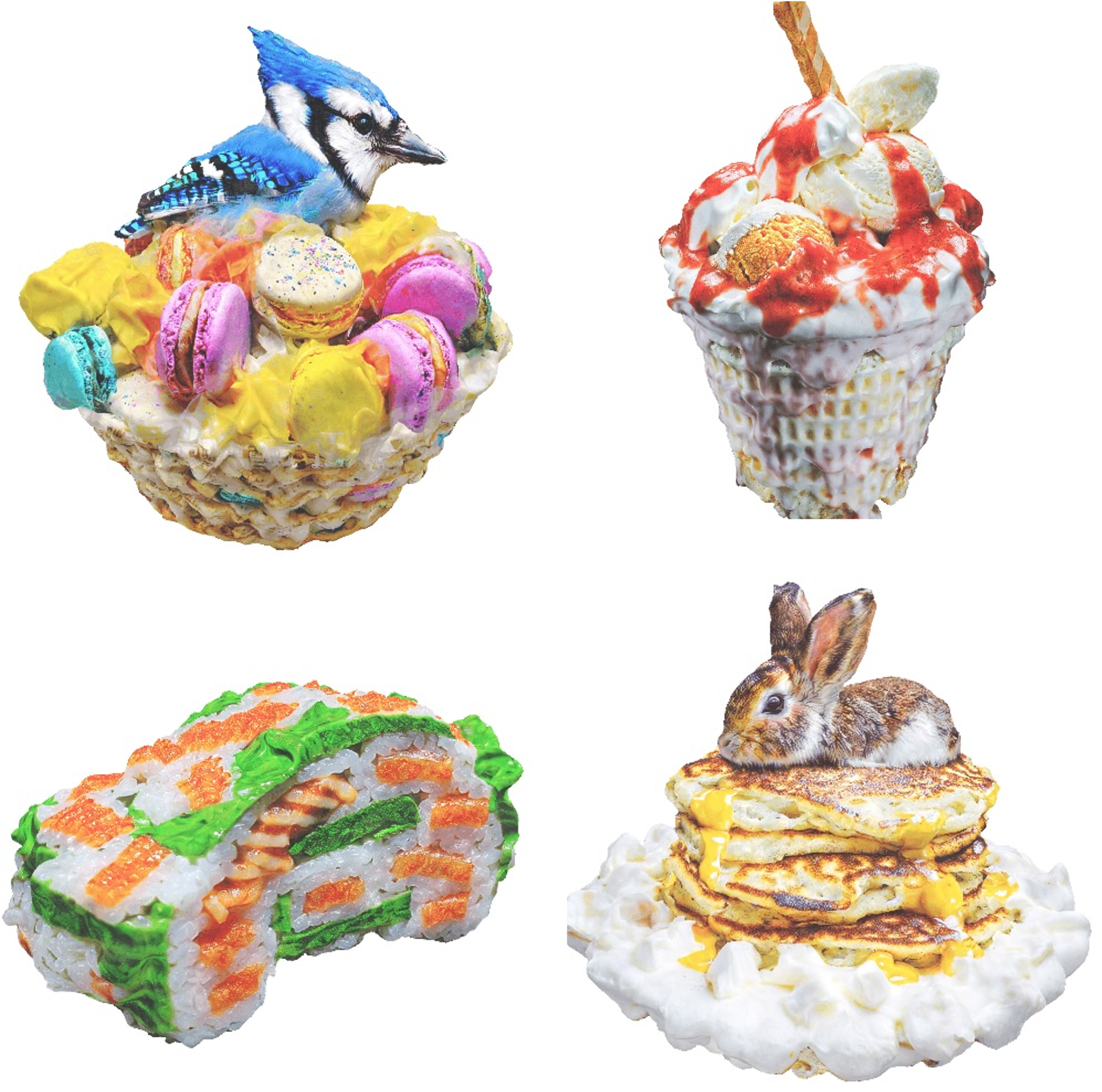}
    \caption{High-quality assets generated by GradeADreamer}
    \label{fig:showcase}
\end{figure}

Apart from SDS \cite{poole2022dreamfusion}, several other score distillation methods have been proposed. For example, Variational Score Distillation (VSD) \cite{wang2023prolificdreamer}, first introduced in ProlificDreamer \cite{wang2023prolificdreamer}, argues that SDS \cite{poole2022dreamfusion} leads to poor quality because it requires a high classifier-free guidance (CFG) \cite{ho2022classifierfree} weight. In contrast, VSD \cite{wang2023prolificdreamer} is compatible with the low CFG \cite{ho2022classifierfree} used in text-to-image generation models. Additionally, Interval Score Matching (ISM) \cite{liang2023luciddreamer} utilizes Denoising Diffusion Implicit Models (DDIM) \cite{song2022denoising} inversion to obtain more consistent pseudo-ground-truths during distillation. Score distillation remains a debated and actively researched subfield. Our results demonstrate that using a simple SDS \cite{poole2022dreamfusion} is sufficient to produce higher quality assets than previously thought as shown in Figure \ref{fig:showcase}. \\

\begin{figure}[ht]
    \centering
    \subfigure[Corgi with multiple faces]{
        \includegraphics[width=0.4\textwidth]{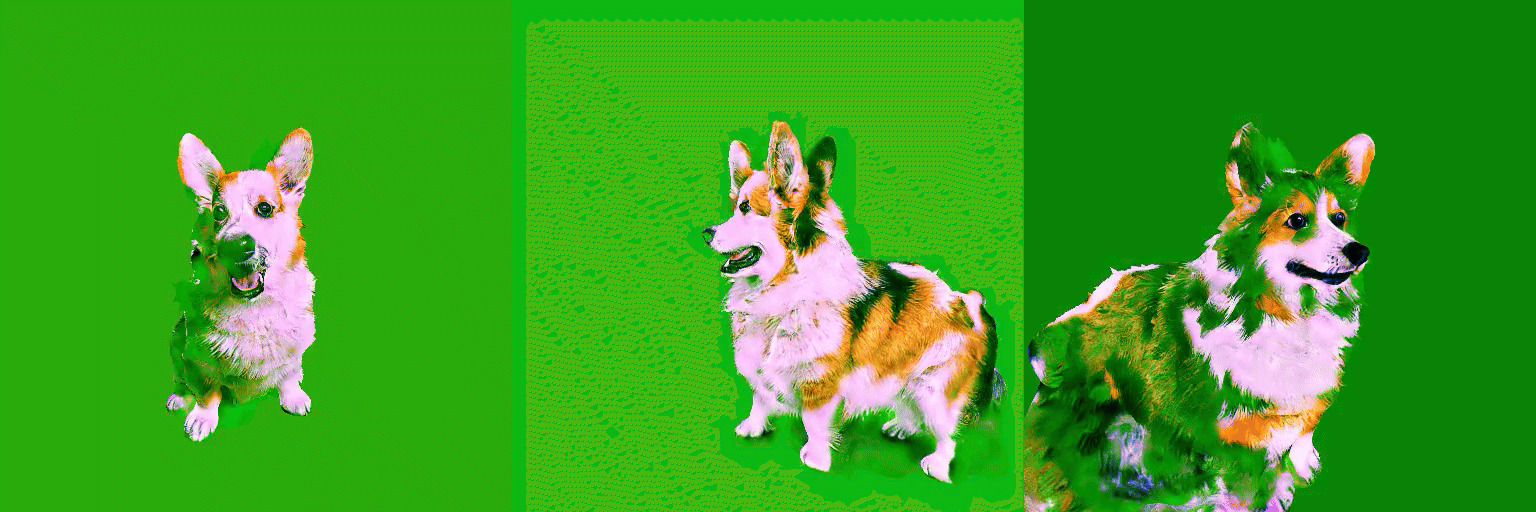}
        \label{fig:image2}
    }
    \hspace{0.05\textwidth} 
    \subfigure[Panda with multiple faces]{
        \includegraphics[width=0.4\textwidth]{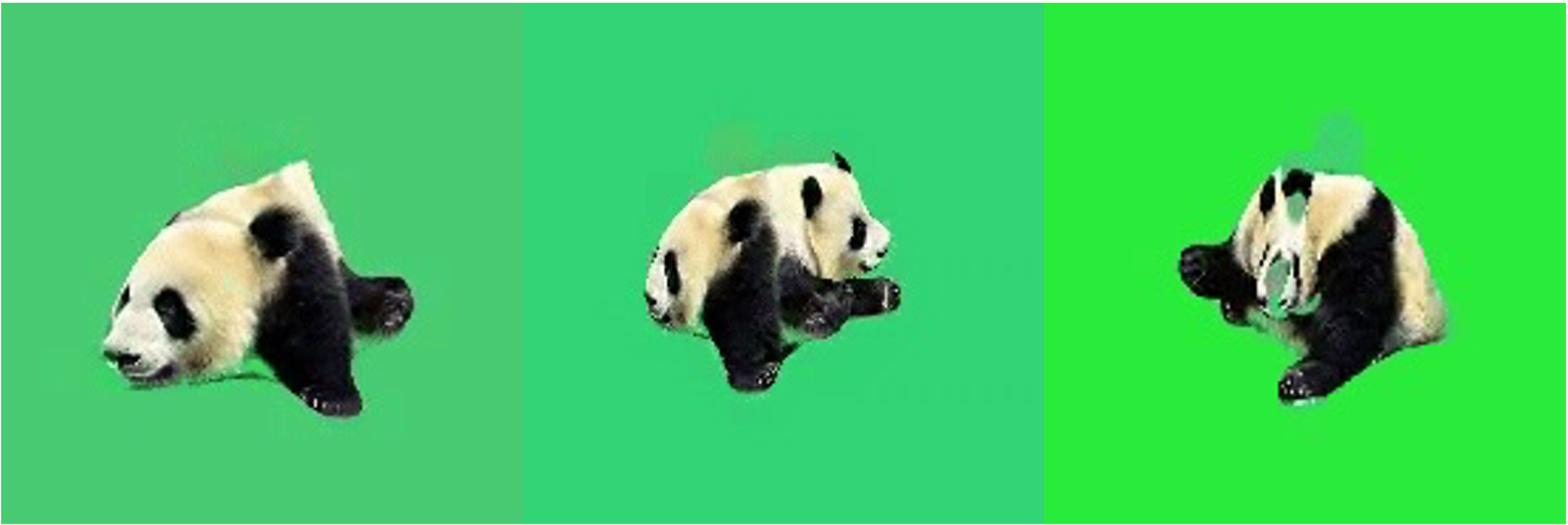}
        \label{fig:image2}
    }
    
    \caption{Examples of Multi-Face Janus Problem \cite{wiki:janus} (Generated with ProlificDreamer \cite{wang2023prolificdreamer})}
    \label{fig:multiface}
\end{figure}

A common problem in text-to-3D generation is the Multi-Face Janus Problem \cite{wiki:janus}, which involves view inconsistency in score-distilling text-to-3D generation, leading to unrealistic and geometrically collapsed 3D objects as illustrated in Figure \ref{fig:multiface}. A Multi-View Diffusion Model, such as MVDream \cite{shi2023MVDream}, addresses this problem by conditioning the generated image from the text-to-image diffusion model with camera poses, thereby improving the model’s viewpoint awareness. This method can significantly reduce the Multi-Face Janus Problem \cite{wiki:janus}. However, despite these advancements, MVDream \cite{shi2023MVDream} often produces poorer styles compared to StableDiffusion \cite{Rombach_2022_CVPR}, due to the limitations of the rendered dataset \cite{shi2023MVDream}. \\

To achieve high-quality 3D asset generation with a low occurrence rate of the Multi-Face Janus Problem \cite{wiki:janus} and low generation time, we propose GradeADreamer, a solution with a pipeline described in three passes. To start, we initialize random 3D Gaussian Splats \cite{kerbl3Dgaussians} and optimize them with the Multi-View Diffusion Model, MVDream \cite{shi2023MVDream}, to establish a base for further optimization. Next, we enter a Gaussian Splats Refinement phase, where we refine the Gaussian Splats \cite{kerbl3Dgaussians} using StableDiffusion \cite{rombach2021highresolution} to achieve better geometrical details. After this, we employ an efficient mesh extraction method described in DreamGaussian \cite{tang2024dreamgaussian}. Finally, we perform a Texture Optimization pass, utilizing the Appearance Refinement technique outlined in Fantasia3D \cite{chen2023fantasia3d}. \\

All passes focus solely on using SDS \cite{poole2022dreamfusion} as the score distillation method to optimize each stage of the 3D content generation pipeline. We leverage Gaussian Splatting \cite{kerbl3Dgaussians} and mesh as our 3D representations for geometry optimization and texture optimization, respectively. \\ \\

In summary, our contributions are as follows:
\begin{enumerate}
    \item We propose GradeADreamer, a high-quality text-to-3D generation model that significantly reduces the occurrence of the Multi-face Janus Problem \cite{wiki:janus} and achieves faster generation time.
    \item We introduce a three-stage optimization process that disentangles geometry and texture optimization while utilizing the Multi-view diffusion model for prior Gaussian Splats \cite{kerbl3Dgaussians} generation.
    \item We evaluate GradeADreamer both qualitatively and quantitatively, demonstrating that it outperforms previous state-of-the-art text-to-3D generation methods \cite{poole2022dreamfusion, lin2023magic3d, wang2023prolificdreamer} in several aspects, including user preference, Multi-face Janus \cite{wiki:janus} occurrence rate, and generation time.
    \item We publicly open-source our text-to-3D generation code and integrate our project in threestudio \cite{threestudio2023}.
\end{enumerate}

\section{Related Works}

\subsection{3D Representations}

In the current state of research for text-to-3D generation, there are multiple 3D Representations being used, the most popular ones are Neural Radiance Fields (NeRF) \cite{mildenhall2020nerf} and Gaussian Splats \cite{kerbl3Dgaussians}. They are particularly useful for this field of research as they are both trainable and they also currently are the most efficient representations to use for that purpose. NeRF trains a neural network \cite{mildenhall2020nerf} whereas Gaussian Splatting optimizes 3D Gaussians \cite{kerbl3Dgaussians}. The current trends exhibit a comparable prevalence of NeRF-based text-to-3D studies \cite{wang2023prolificdreamer, zhang2023text2nerf, liu2024direct, li2024grounded, kant2024spad} and those based on Gaussian Splatting  \cite{tang2024dreamgaussian, yi2024mvgamba, li2024controllable, miao2024pla4d, yi2023gaussiandreamer}.\\

The differences between NeRF \cite{mildenhall2020nerf}, Gaussian Splats \cite{kerbl3Dgaussians}, and other more traditional 3D representations lie in their optimization capabilities. Techniques such as photogrammetry are not efficient for these tasks. NeRF \cite{mildenhall2020nerf} and Gaussian Splats \cite{kerbl3Dgaussians}, on the other hand, are specifically designed to be optimized in the tasks such as 3D rendering and reconstruction, offering more advanced and efficient solutions for generating detailed 3D models from textual descriptions.\\

The selection of NeRF \cite{mildenhall2020nerf} or Gaussian Splatting \cite{kerbl3Dgaussians} for text-to-3D generation depends largely on the specific requirements of the research task. NeRF typically yields higher-quality results but requires more time to converge compared to Gaussian Splatting methods \cite{chen2024textto3d, tang2024dreamgaussian}. This trade-off between quality and convergence time underscores the importance of choosing the appropriate 3D generation technique based on the specific computational resources and performance criteria of the research objectives.

\subsection{Score Distillation Methods}

Due to the limited availability of 3D data for training text-to-3D models, Score Distillation Methods are widely used to effectively distill knowledge from pre-trained text-to-image diffusion models, such as StableDiffusion \cite{rombach2021highresolution}. This approach leverages the strengths of these well-trained text-to-image models to achieve state-of-the-art performance in text-to-3D generation. By transferring the expertise from 2D to 3D, Score Distillation Methods enhance the capability to generate accurate and detailed 3D models from textual descriptions, overcoming the data scarcity challenge in the 3D domain. \\

Score Distillation Sampling (SDS) \cite{poole2022dreamfusion} and Variational Score Distillation (VSD) \cite{wang2023prolificdreamer} have demonstrated significant promise in recent state-of-the-art advancements in text-to-3D generation. While SDS and VSD share this foundational approach, VSD sets itself apart from SDS by employing particle-based variational inference \cite{chen2018unified, dong2023particlebased, liu2019stein}. This technique is perceived to offer advantages by mitigating issues such as oversaturation and oversmoothing. However, our findings show that a well-designed pipeline using SDS alone can effectively address these challenges, demonstrating that SDS can achieve comparable performance without the additional complexity introduced by VSD.

\subsection{Mesh Extraction from Gaussian Splatting}

Gaussian Splatting \cite{kerbl3Dgaussians} represents 3D space using Gaussian Splats, which are not directly compatible with the traditional 3D mesh formats commonly employed in industry. Consequently, additional steps are required to convert these splats into conventional mesh representations suitable for practical applications. \\

For this task, studies such as SuGaR \cite{guedon2023sugar} initially utilize a traditional mesh extraction method as outlined in the original Gaussian Splatting paper \cite{kerbl3Dgaussians}. Subsequently, they refine this approach by employing a hybrid representation of meshes and Gaussian Splats. This refinement enhances accuracy and smoothness, enabling the achievement of state-of-the-art results. \\

Most of the research done on mesh extraction is based not just on Gaussian Splats \cite{kerbl3Dgaussians} as input but also on images \cite{guedon2023sugar, chen2024pgsr, fan2024trim, shen2024gamba} from which the mesh should be extracted from. A subtask that is not explored enough would be efficient mesh extraction from Gaussian Splats alone.

\section{Approach}

\begin{figure*}[t]
    \centering
    \includegraphics[width=\linewidth]{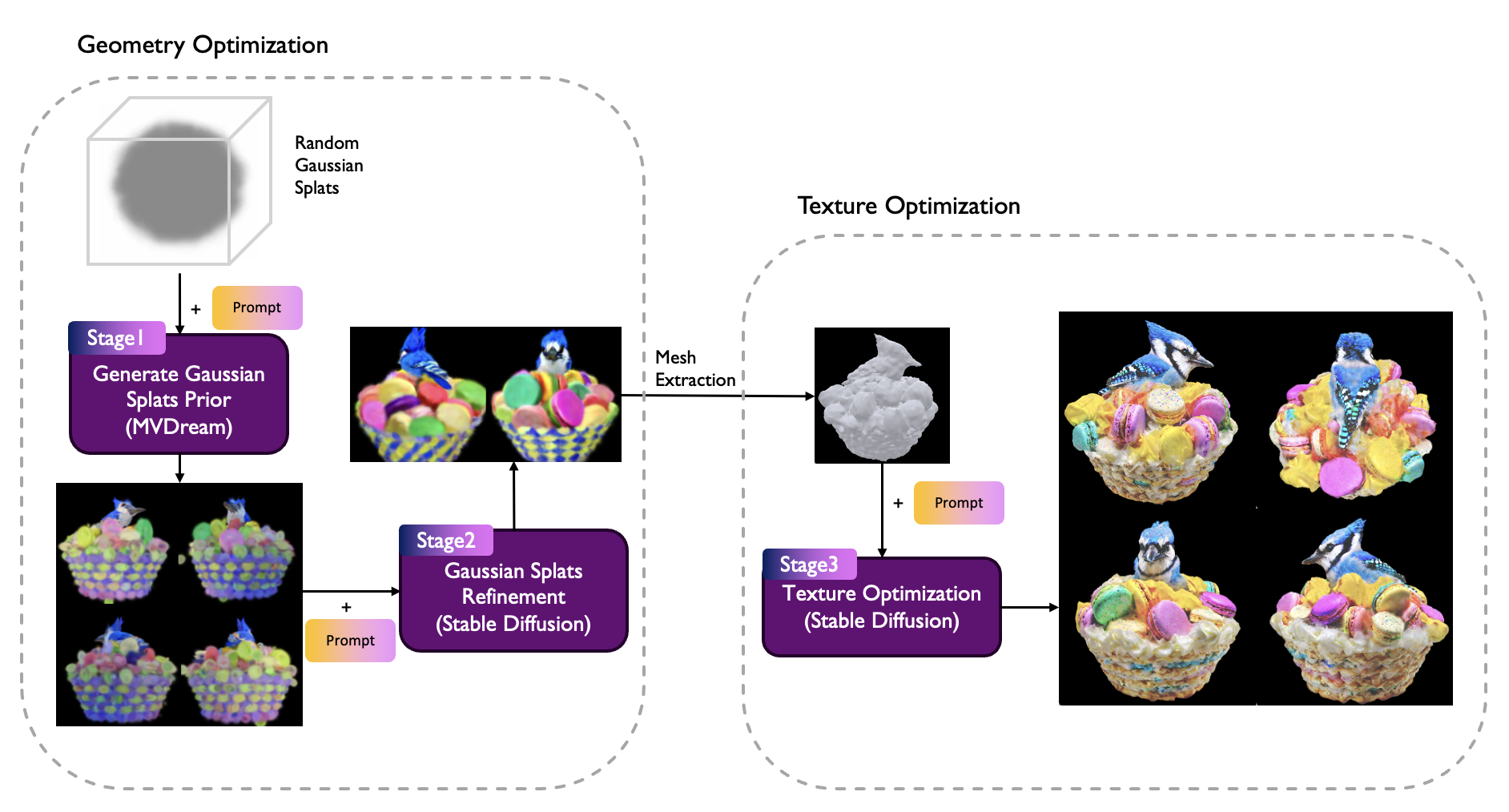}
    \caption{\textbf{Overview of GradeADreamer}. The proposed method consists of three optimization steps. The first step involves optimizing random Gaussian Splats using MVDream to obtain a Gaussian Splats prior (see Section \ref{approach:stage1}). In the second step, this prior is refined using StableDiffusion (see Section \ref{approach:stage2}). Finally, the third step employs texture optimization on a mesh, guided by StableDiffusion (see Section \ref{approach:stage3}).}
    \label{fig:pipeline}
\end{figure*}

We introduce a novel text-to-3D training pipeline designed to maintain low generation time while mitigating the Multi-face Janus Problem \cite{wiki:janus}. Our pipeline is composed of three distinct stages:
\begin{enumerate}
    \item Gaussian Splats Prior Generation
    \item Gaussian Splats Refinement
    \item Texture Optimization
\end{enumerate}

The training pipeline is executed sequentially. We follow the approach of Fantasia3D \cite{chen2023fantasia3d} in disentangling geometry and texture optimization. The first two stages focus on geometry optimization, while the last stage is responsible for texture optimization. However, unlike Fantasia3D \cite{chen2023fantasia3d}, which requires a handcrafted mesh for initialization, our pipeline can produce 3D models without any mesh prior. An overview of our proposed pipeline is shown in Figure \ref{fig:pipeline}.

\subsection{3D Representation Choices}

For the first and second stages, we utilize Gaussian Splatting \cite{kerbl3Dgaussians} as our 3D representation. This choice is based on its significantly faster convergence compared to Neural Radiance Fields (NeRF) \cite{mildenhall2020nerf} for 3D generation tasks, as demonstrated by DreamGaussian \cite{tang2024dreamgaussian}. To achieve high-quality textured meshes, we extract meshes from the Gaussian Splats \cite{kerbl3Dgaussians} and then proceed with texture refinement during the later stages of our pipeline.

\subsection{Training Stages}

\subsubsection{Gaussian Splats Prior Generation}
\label{approach:stage1}

Starting with random Gaussian Splats \cite{kerbl3Dgaussians}, we employ MVDream \cite{shi2023MVDream}, a 2D multi-view diffusion model, as our guidance model to optimize the random Gaussian Splats \cite{kerbl3Dgaussians} into a Gaussian Prior with coarse details of the result. This approach effectively reduces the likelihood of encountering the Multi-face Janus Problem \cite{wiki:janus}.

\subsubsection{Gaussian Splats Refinement}
\label{approach:stage2}

In this stage, we refine the Gaussian Splats \cite{kerbl3Dgaussians} to enhance the geometry using StableDiffusion 2.1 \cite{Rombach_2022_CVPR} as our guidance. This step is crucial because MVDream \cite{shi2023MVDream} can only generate low-resolution images (256x256), whereas the StableDiffusion 2.1 \cite{Rombach_2022_CVPR} can produce higher resolution outputs (512x512). We also observe, similar to MVDream \cite{shi2023MVDream}, that the StableDiffusion 2.1 \cite{Rombach_2022_CVPR} yields better generation styles, likely due to the influence of the rendered training dataset on MVDream \cite{shi2023MVDream} outputs.

Following the Gaussian Refinement stage, we extract the mesh using an efficient mesh extraction algorithm as described by DreamGaussian \cite{tang2024dreamgaussian}.

\subsubsection{Texture Optimization}
\label{approach:stage3}

In the final stage, we optimize the spatially varying Physically-Based Rendering (PBR) material model on the mesh. Specifically, we focus on three distinct components of the material model: the diffuse map \( k_d \in \mathbb{R}^3 \), which represents the inherent color of the material without any lighting effects; the roughness and metallic term \( k_{rm} \in \mathbb{R}^2 \), which determines the micro-surface detail and metalness of the material; and the normal variation term \( k_{n} \in \mathbb{R}^3 \), which details surface texture by perturbing the normal vectors of the surface. For this step, we follow the techniques outlined in Fantasia3D \cite{chen2023fantasia3d} to achieve high-quality, detailed textures.

\subsection{Score Distillation Choices}

We examine the differences between Score Distillation Sampling (SDS) proposed in Dreamfusion \cite{poole2022dreamfusion} and Variational Score Distillation (VSD) proposed in ProlificDreamer \cite{wang2023prolificdreamer}. Our observations are as follows:

\begin{itemize}
    \item \textbf{Training Time} : We observe that using Variational Score Distillation (VSD) \cite{wang2023prolificdreamer} extends the training time compared to using SDS \cite{poole2022dreamfusion}.
    \item \textbf{Quality of Results} : Our results demonstrate that using SDS \cite{poole2022dreamfusion} is sufficient for generating high-quality 3D models in text-to-3D generation tasks.
\end{itemize}

Based on these findings, we employ SDS \cite{poole2022dreamfusion} as our objective for score distillation in all three stages of our pipeline. 

The gradient of the SDS loss, as described in \cite{poole2022dreamfusion}, can be expressed by the following equation:

\begin{equation}
    \nabla_\theta \mathcal{L}_{\text{SDS}} = \mathbb{E}_{\epsilon, t} \left[ w(t) (\epsilon_\phi(x_t | y, t) - \epsilon) \frac{\partial x}{\partial \theta}\right]
    \label{eq:sds}
\end{equation}

where $\theta$ represents the parameters of a 3D representation, $x$ is an image rendered from this 3D representation through a transformation function $g$ expressed as $x=g(\theta)$. $x_t$is a residual noise image predicted by a pretrained 2D diffusion model $\epsilon_\phi$ conditioned on a prompt embedding $y$ and a timestep $t$. And $\epsilon$ is a noisy image drawn from a normal distribution, $\epsilon \sim \mathcal{N}(0, I)$.

\section{Experiments}

In this section, we detail our experiment implementation and validate the proposed approach's effectiveness through both quantitative and qualitative analyses by comparing GradeADreamer to previous state-of-the-art methods in text-to-3D generation.

\subsection{Implementation Details}

\subsubsection{Training Resources}

Our entire pipeline can be run on a single RTX 3090 GPU, with a generation time of only 30 minutes per 3D asset. In stage 1, it takes 12 GB of GPU RAM and 4 minutes; stage 2 requires 15 GB and 6 minutes; and stage 3 uses 16 GB and 16 minutes. The remaining 4 minutes are allocated for mesh extraction, Gaussian Splats \cite{kerbl3Dgaussians} export, and the final result validation step.

\subsubsection{Hyperparameters}

\paragraph{Initialization} We initialize 6000 points of Gaussian Splats \cite{kerbl3Dgaussians}. We’ve found that using a large number of initial Gaussian Splats \cite{kerbl3Dgaussians} extends the optimization time and may not converge in some cases. 

\paragraph{Gaussian Splats Prior Generation} In the first stage, similar to Dreamtime \cite{huang2024dreamtime}, we employ a linear annealing timestep from 0.98 to 0.02 for the entire 700 optimization steps. We’ve found that applying the linear annealing timestep yields better results in geometry. The densification and pruning interval is set to 55. Following DreamGaussian \cite{tang2024dreamgaussian}, the gradient threshold for densification is set to 0.01, and we perform an opacity reset at step 500. In this stage, we use a batch size of 1, consisting of 4 image views as proposed in MVDream \cite{shi2023MVDream}.

\paragraph{Gaussian Splats Refinement} In the second stage, there are 700 optimization steps. We utilize a linear annealing timestep from 0.98 to 0.02 for the first 200 steps. For steps 200-300, we uniformly sample the timestep from the range 0.02 to 0.98, $t \sim \mathcal{U}(0.02, 0.98)$, and we anneal it to $t \sim \mathcal{U}(0.02, 0.50)$ for the last 400 steps to focus on distilling elaborate details, following ProlificDreamer \cite{wang2023prolificdreamer}. We set the densification interval to 50 and the opacity reset interval to 300. The batch size is set to 2, as we observe that it converges faster than a batch size of 1 and yields a reasonable geometric appearance.

\paragraph{Gaussian Training} In the first two stages, we use a linear decaying positional learning rate for Gaussian Splatting from 0.001 to $1.6 \times 10^{-6}$ over 300 steps, and we set the feature and opacity learning rates to 0.005 and 0.05, respectively.

\paragraph{Texture Optimization} For the last stage, we follow the hyperparameters provided by Fantasia3D \cite{chen2023fantasia3d}, optimizing for 2000 iterations with a batch size of 4 using an SDS \cite{poole2022dreamfusion} weight strategy equal to 1. We employ SDS \cite{poole2022dreamfusion} as a score distillation objective for all stages with a classifier-free guidance (CFG \cite{ho2022classifierfree}) value of 100.

\subsection{Baselines}

We utilize three previous state-of-the-art text-to-3D generation models as our baselines: DreamFusion \cite{poole2022dreamfusion}, Magic3D \cite{lin2023magic3d}, and ProlificDreamer \cite{wang2023prolificdreamer}. Our results will be compared to these baselines both qualitatively and quantitatively in the following sections.

\subsection{Qualitative Evaluations}

\begin{figure*}[htbp]
    \centering
    \includegraphics[width=\linewidth]{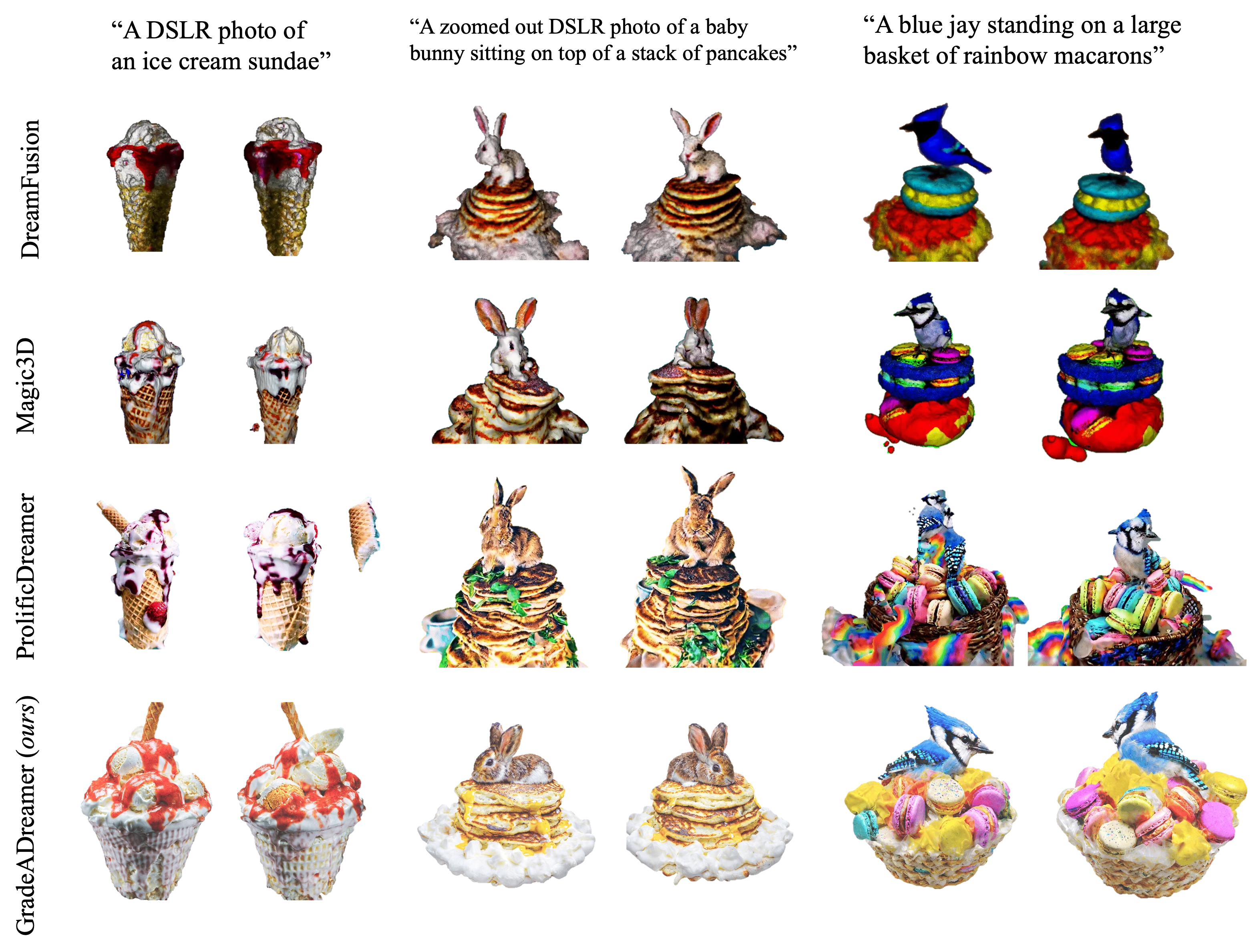}
    \caption{Qualitative results}
    \label{fig:qualitative_results}
\end{figure*}

We qualitatively assess the performance of our proposed method, GradeADreamer, against the established baselines. As illustrated in Figure \ref{fig:qualitative_results}, our results significantly improve the geometric features of the generated assets, addressing the geometry flaws caused by the Multi-face Janus Problem \cite{wiki:janus} that other baselines \cite{poole2022dreamfusion, lin2023magic3d, wang2023prolificdreamer} encounter. Additionally, our method produces high-frequency, high-quality textures, whereas DreamFusion \cite{poole2022dreamfusion} and Magic3D \cite{lin2023magic3d} struggle with issues of over-saturation and over-smoothing. For more qualitative results, please refer to the Appendix \ref{appendix:qualitative}.

\subsection{Quantitative Evaluations}

\subsubsection{User Study}

\begin{figure}[htbp]
    \centering
    \includegraphics[width=0.4\textwidth]{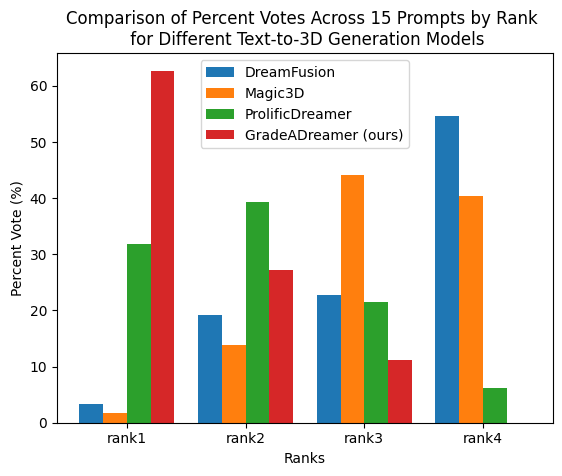}
    \caption{Rank distribution comparison from user study}
    \label{fig:userstudy}
\end{figure}

We conducted a user study to evaluate our text-to-3D model against other baselines \cite{poole2022dreamfusion, lin2023magic3d, wang2023prolificdreamer}. The study involved 54 participants who were asked to rank 3D models generated from 15 different prompts. These results were non-cherry-picked, meaning all data was selected without selectively showcasing only the best outcomes. This approach ensures a fair comparison and avoids the bias and misleading outcomes that come from cherry-picking. Participants ranked the results from best to worst for each prompt. As shown in Figure \ref{fig:userstudy}, Our model decisively outperformed the others, achieving a significantly higher win rate for the top rank compared to the baselines \cite{poole2022dreamfusion, lin2023magic3d, wang2023prolificdreamer}. Additionally, the average rankings presented in Table \ref{table:averageranking} demonstrate that our model gets the highest average ranking among the four selected methods. This study shows that the 3D models created by our pipeline were consistently preferred by users.

\begin{table}[H]
    \begin{center}
    \begin{tabular}{lc}
    \hline
    Name & Average Ranking ($\downarrow$) \\
    \hline
    DreamFusion & 3.29 \\
    Magic3D & 3.23 \\
    ProlificDreamer & 2.02 \\
    \textbf{GradeADreamer (\textit{ours})} & \textbf{1.49} \\
    \hline
    \end{tabular}
    \end{center}
    \caption{Average ranking from the user study}
    \label{table:averageranking}
\end{table}

\subsubsection{3D-FID Score}

We also performed an additional evaluation method utilizing a variant of the Fréchet Inception Distance (FID) score. The FID score is a widely used metric for comparing two datasets, specifically quantifying how well generated content captures the statistical properties of the original content. This metric is predominantly used in the context of 2D image generation, where it requires the generated images and a set of ground truth images or approximate target results for comparison.

In our study, GradeADreamer operates on 3D generated content, necessitating certain modifications to the standard FID calculation. To address this, we employed a 3D-FID score. The process for calculating the 3D-FID score involves generating a reference image using StableDiffusion \cite{rombach2021highresolution}, followed by calculating a 2D-FID score for a series of multiple views of the 3D generated model. Specifically, we used 10 different views by rotating the model along the yaw axis. The final 3D-FID score is obtained by averaging the scores from each of these views.

In our comparison table of 3D-FID scores (see Table \ref{table:3dFID}), we computed the average score from 15 randomly generated content samples. Each Text-to-3D model was tasked with generating the same content, and the 3D-FID score was calculated for each model, followed by averaging the scores from the 15 samples. This rigorous evaluation demonstrated the robustness of GradeADreamer, as evidenced by its 3D-FID score.

\begin{table}[H]
    \begin{center}
    \begin{tabular}{lc}
    \hline
    Name & 3D-FID ($\downarrow$) \\
    \hline
    DreamFusion & 68.41 \\
    Magic3D & 75.96 \\
    ProlificDreamer & 58.75 \\
    \textbf{GradeADreamer (\textit{ours})} & \textbf{47.69} \\
    \hline
    \end{tabular}
    \end{center}
    \caption{3D-FID Score}
    \label{table:3dFID}
\end{table}

\subsubsection{Multi-face Janus Problem}

We measured the occurrence in percentage of the Multi-face Janus Problem \cite{wiki:janus} over 15 prompts using our model, compared to other baselines \cite{poole2022dreamfusion, lin2023magic3d, wang2023prolificdreamer}. As shown in Table \ref{table:janus}, by using MVDream \cite{shi2023MVDream} as a guidance model for the first stage, we achieved a significantly lower occurrence rate of the Multi-face Janus Problem \cite{wiki:janus}, approximately 6 to 7 times lower than the baselines \cite{poole2022dreamfusion, lin2023magic3d, wang2023prolificdreamer}.

\begin{table}[H]
    \begin{center}
    \begin{tabular}{lc}
    \hline
    Name & \% Multi-face ($\downarrow$) \\
    \hline
    DreamFusion & 35.71 \\
    Magic3D & 35.71 \\
    ProlificDreamer & 40.00 \\
    \textbf{GradeADreamer (\textit{ours})} & \textbf{6.67} \\
    \hline
    \end{tabular}
    \end{center}
    \caption{Percentage of Multi-face Janus problem \cite{wiki:janus} occurrence}
    \label{table:janus}
\end{table}

\subsubsection{Runtime}

We evaluated the generation time per prompt for the selected four methods to assess the efficiency of our proposed method. The results are summarized in Table \ref{table:runtime}.

Our model achieves a significant reduction in generation time compared to Magic3D \cite{lin2023magic3d} and ProlificDreamer \cite{wang2023prolificdreamer}, making it highly suitable for applications requiring quick asset generation. The equivalence in speed to DreamFusion \cite{poole2022dreamfusion}, one of the fastest models, highlights the effectiveness of our pipeline without compromising on the quality of the generated assets, as proven in Table \ref{table:averageranking} and Table \ref{table:3dFID}. This efficiency is crucial for the applications and scenarios where rapid prototyping and iteration are necessary.

\begin{table}[H]
    \begin{center}
    \begin{tabular}{lc}
    \hline
    Name & Generation Time ($\downarrow$) \\
    \hline
    \textbf{DreamFusion} & \textbf{30 mins} \\
    Magic3D & 1 hour \\
    ProlificDreamer & 10 hours \\
    \textbf{GradeADreamer (\textit{ours})} & \textbf{30 mins} \\
    \hline
    \end{tabular}
    \end{center}
    \caption{Generation time per prompt}
    \label{table:runtime}
\end{table}

\section{Conclusion}

In this paper, we present GradeADreamer, a novel text-to-3D generation pipeline designed to produce high-quality 3D assets while addressing the Multi-face Janus Problem \cite{wiki:janus} and maintaining low generation time. Our pipeline comprises three stages: Gaussian Splats Prior Generation, Gaussian Splats Refinement, and Texture Optimization. The first two stages leverage Gaussian Splatting as a 3D representation to optimize geometry, while the third stage focuses on texture optimization using Mesh representation to ensure high-quality textures.

To mitigate the Multi-face Janus Problem \cite{wiki:janus}, we employ a Multi-view Diffusion model, MVDream \cite{shi2023MVDream}, in the initial stage to generate priors, and subsequently use StableDiffusion \cite{Rombach_2022_CVPR} for enhanced generation quality. Our experimental results demonstrate that GradeADreamer achieves the highest average ranking in user studies and the lowest generation time, around 30 minutes per asset, compared to all baseline models. Additionally, the occurrence rate of the Multi-face Janus Problem \cite{wiki:janus} significantly decreased.

\paragraph{Limitations} Despite these advancements, GradeADreamer has some limitations. For example, the performance is constrained by the StableDiffusion \cite{Rombach_2022_CVPR} Text Encoder, and there are still instances of the Multi-face problem since we still use StableDiffusion \cite{Rombach_2022_CVPR} as guidance for the last two stages. Furthermore, there is potential for improvement in the mesh extraction method from Gaussian Splats to obtain higher quality mesh.

In summary, GradeADreamer represents a significant step forward in the field of text-to-3D generation, combining efficiency with high-quality output and effectively addressing key challenges in the domain. Future work will focus on overcoming the current limitations and further enhancing the performance and quality of the generated assets.

\section{Contribution}

\paragraph{Trapoom Ukarapol} Propose the main idea of GradeADreamer, implementing the pipeline, generate results, qualitative evaluation, Multi-face Janus Evaluation, user study, presentation and paper work.

\paragraph{Kevin Pruvost} Mesh extraction experiments, geometry optimization experiments, generate results, 3D-FID evaluation, user study, presentation and paper work.

\section{Acknowledgement}

We would like to express our sincere gratitude to the Deep Learning 2024 Spring course at Tsinghua University for providing us with a robust foundational knowledge of deep learning and offering us the opportunity to undertake a project aligned with our interests. We are particularly grateful to Professor Xiaolin Hu and Professor Jun Zhu, the course lecturers, for their invaluable guidance. Additionally, we extend our appreciation to Professor Hualiu Ping and Professor Liu Yongjin for supplying the necessary computational resources. Finally, we would like to thank the 54 participants in our user study for their valuable contributions.

{\small
\bibliographystyle{ieee_fullname}
\bibliography{main}
}

\newpage 

\appendix

\section{Evaluation Prompts}

We selected 15 commonly used prompts for text-to-3D generation from several research papers to ensure fair evaluations. The prompts are listed in Table \ref{table:prompt}

\begin{table}[H]
    \centering
    \begin{tabular}{ll}
    \hline
    \textbf{\#} & \textbf{Prompt} \\
    \hline
    1 & A DSLR photo of a panda \\
    \hline
    2 & A high quality photo of a furry corgi \\
    \hline
    3 & \makecell[l]{A blue jay standing on a \\ large basket of rainbow macarons} \\
    \hline
    4 & A ripe strawberry \\
    \hline
    5 & A DSLR photo of an ice cream sundae \\
    \hline
    6 & A DSLR photo of car made out of sushi \\
    \hline
    7 & A pineapple \\
    \hline
    8 & A DSLR photo of an astronaut is riding a horse \\
    \hline
    9 & A model of a house in Tudor style \\
    \hline
    10 & A highly detailed sand castle \\
    \hline
    11 & A plate piled high with chocolate chip cookies \\
    \hline
    12 & \makecell[l]{A zoomed out DSLR photo of a baby bunny sitting \\ on top of a stack of pancakes} \\
    \hline
    13 & A photo of a horse walking \\
    \hline
    14 & \makecell[l]{A DSLR photo of a plate of fried chicken and \\ waffles with maple syrup on them} \\
    \hline
    15 & A bulldog wearing a black pirate hat \\
    \hline
    \end{tabular}
    \caption{Prompts used for Evaluations}
    \label{table:prompt}
\end{table}

\section{Additional Qualitative Results}
\label{appendix:qualitative}
\begin{figure*}[htbp]
    \centering
    \includegraphics[width=\linewidth]{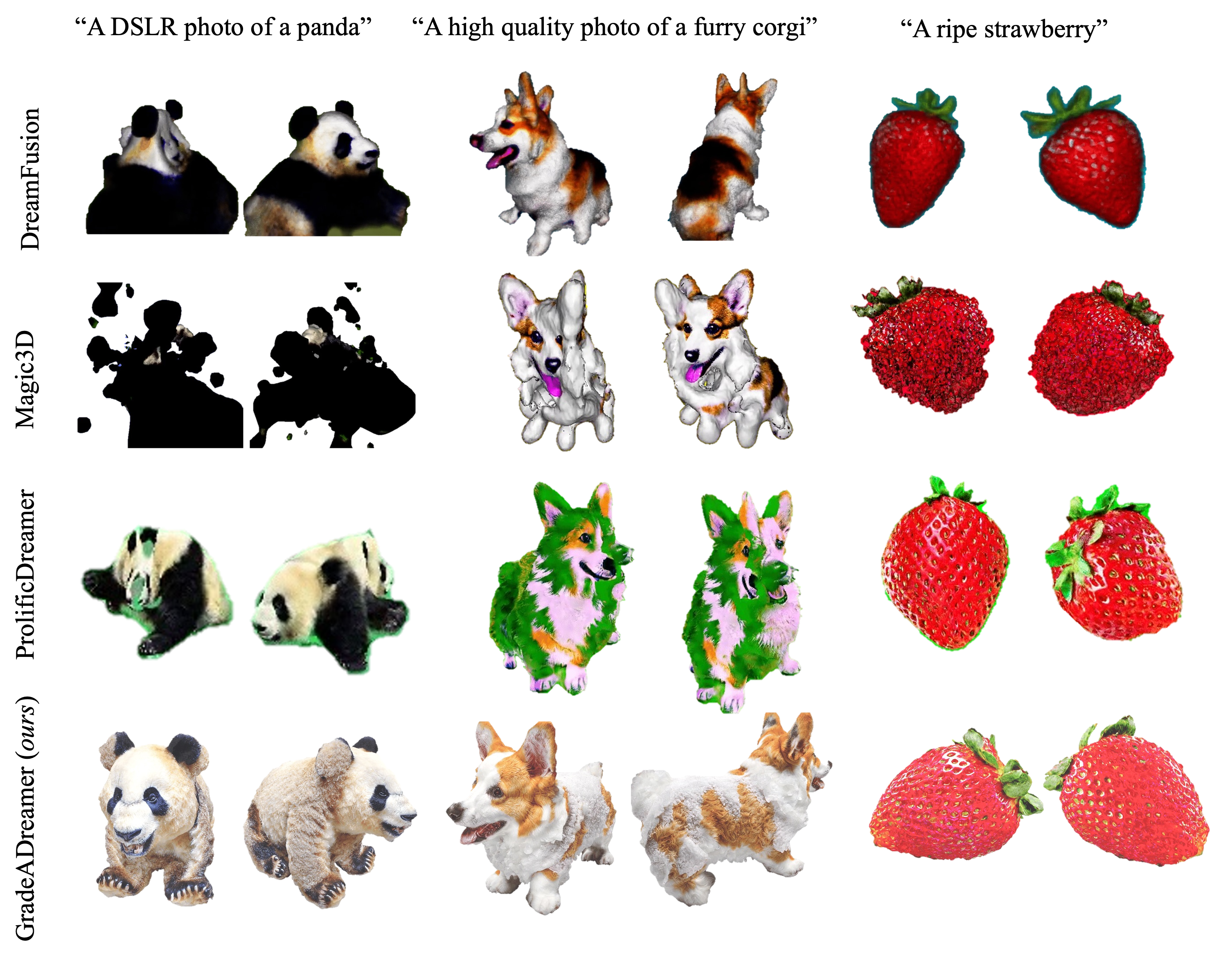}
    \caption{More qualitative results}
    \label{fig:qualitative_results_1}
\end{figure*}

\begin{figure*}[htbp]
    \centering
    \includegraphics[width=\linewidth]{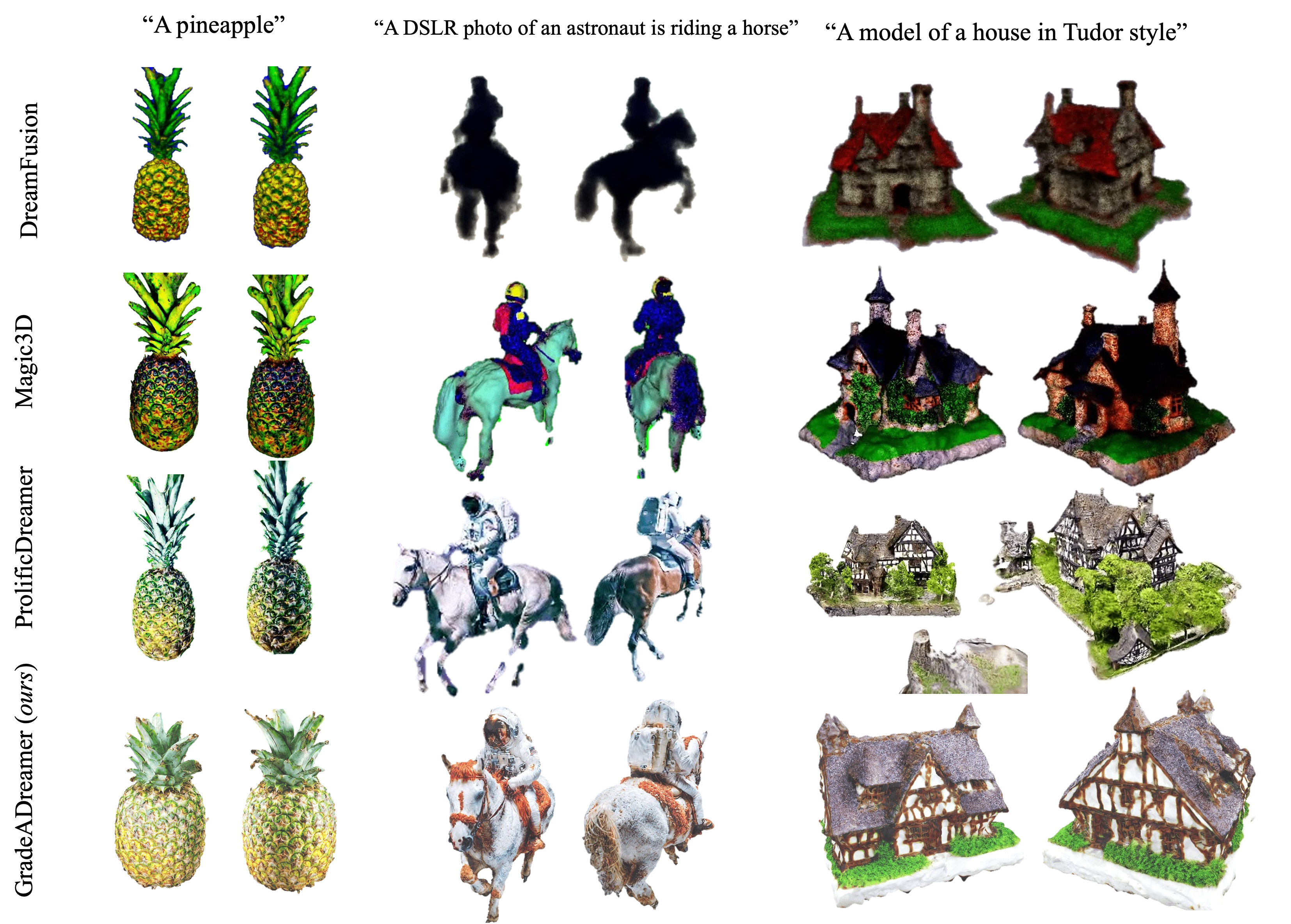}
    \caption{More qualitative results}
    \label{fig:qualitative_results_2}
\end{figure*}

\begin{figure*}[htbp]
    \centering
    \includegraphics[width=\linewidth]{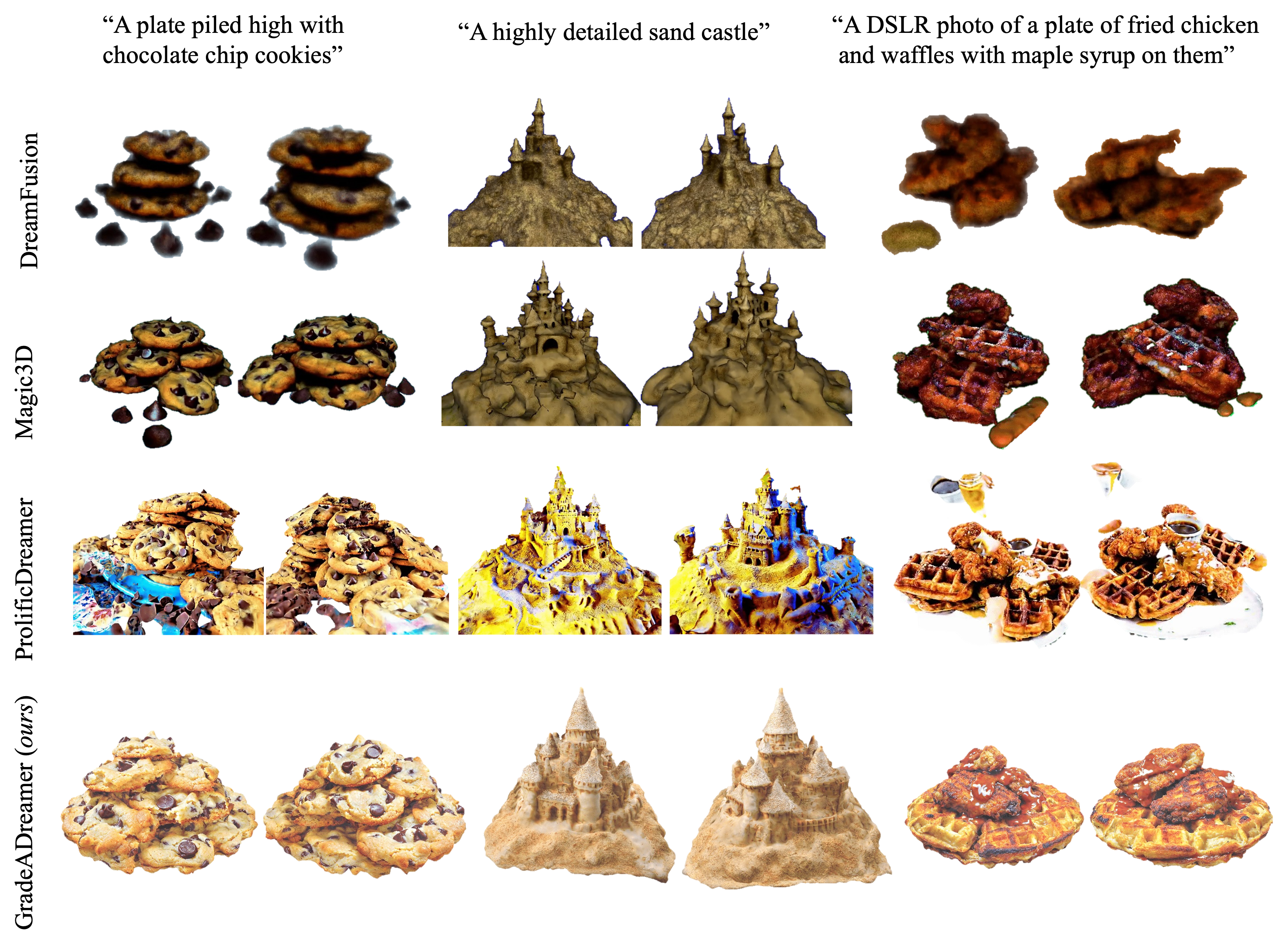}
    \caption{More qualitative results}
    \label{fig:qualitative_results_3}
\end{figure*}

We provide more text-to-3D qualitative comparisons of our proposed method against the other baselines in Figure \ref{fig:qualitative_results_1}, \ref{fig:qualitative_results_2}, and \ref{fig:qualitative_results_3}.

\newpage 

\end{document}